%% file: iclr2023_conference_tinypaper.tex
\documentclass{article} 
\usepackage{iclr2023_conference_tinypaper,times}
\usepackage{graphicx}

\input{math_commands.tex}

\usepackage{comment}
\usepackage{hyperref}
\usepackage{url}
\usepackage{algorithm,  algpseudocode}
\usepackage{multirow}

\title{Region Mixup 
}

\author{Saptarshi Saha \& Utpal Garain  \\ 
Indian Statistical Institute Kolkata  \\
\texttt{\{s.saha\_r,utpal\}@isical.ac.in} \\
}

\iclrfinalcopy 
\begin{document}

\maketitle

\begin{abstract}
This paper introduces a simple extension of mixup \citep{zhang2018mixup} data augmentation to enhance generalization in visual recognition tasks. Unlike the vanilla mixup method, which blends entire images, our approach focuses on combining regions from multiple images.
\end{abstract}

\section{Introduction}
Mixup \citep{zhang2018mixup} is a data augmentation method that trains models on weighted averages of randomly paired training points. The averaging weights are typically sampled from a beta distribution with parameter $\alpha$, where $\alpha$ ensures that the generated training set remains close to the original dataset.
Mixup-generated perturbations may adhere only to the direction towards any arbitrary data point, potentially resulting in suboptimal regularization \citep{DBLP:conf/aaai/GuoMZ19}. To this end, we propose Region Mixup, an approach emphasizing the integration of regions from multiple images. While various mixup variants \citep{Verma2018ManifoldMB,kim2021comixup,Liu2021AutoMixUT} have been proposed to address suboptimal regularization, including those considering convex combinations of more than two points, yet none explicitly strive to interpolate at the level of regions. Closest to our work is
CutMix \citep{Yun2019CutMixRS}. However, CutMix does not interpolate regions; instead, it cuts and pastes patches between training images.

\section{Region Mixup}

Let $ x \in \mathbb{R}^{W \times H \times C}$ and $y$ represent a training image and its corresponding label, respectively. The objective of region mixup is to create a new training sample $(\tilde{x}, \tilde{y})$ by combining regions from multiple training samples $(x_{A}, y_{A}), (x_{B_{1}}, y_{B_{1}}), (x_{B_{2}}, y_{B_{2}}), \ldots, (x_{B_{k^{2}}}, y_{B_{k^{2}}})$. The combining operation is defined as follows:
\begin{align}\label{Mixup_eq}
    \tilde{x}  = \sum_{j=1}^{k^{2}} \lambda_{j} M_{j} \odot x_{A} + (1-\lambda_{j}) M_{j} \odot x_{B_{j}}, \quad  \text{and} \qquad 
    \tilde{y}  = \dfrac{1}{k^{2}} \sum_{j=1}^{k^{2}} \lambda_{j}  y_{A} + (1-\lambda_{j})  y_{B_{j}},
\end{align}
where $M_{j}\in \{0,1\}^{W\times H}$ denotes a binary mask representing the region to be mixed up from two images $x_{A}$ and $x_{B_{j}}$, and $\sum_{j=1}^{k^{2}}M_{j}=\mathbf{1}$. The operation 
$\odot$ denotes element-wise multiplication. If
$k = 1$, we recover standard mixup regularization. 

\begin{algorithm}[H]
\caption{Region Mixup at $t$-th training iteration }
\begin{algorithmic}[1]
\Statex \textbf{Input:} Mini-batch  ($\mathbf{x}$, $\mathbf{y}$), classifier $f$ with parameters $\theta_{t-1}$, and model optimizer SGD 
\State Sample mixup parameters $\lambda_{1},\lambda_{2},...,\lambda_{k^{2}}\sim Beta(\alpha,\alpha) $
\State $(\mathbf{x}_{A}, \mathbf{y}_{A}) = (\mathbf{x}, \mathbf{y})$ ;  $(\mathbf{x}_{B_{j}}, \mathbf{y}_{B_{j}}) $ = \textbf{RandomPermute}($\mathbf{x}, \mathbf{y}$)  for $j=1$ to $k^{2}$ 
\State Compute $(\tilde{\mathbf{x}}, \tilde{\mathbf{y}})$ using equation \ref{Mixup_eq}
\State $$\mathcal{L}= \textcolor{magenta}{\textit{CE}\Big(f(\mathbf{x}_{A}), \mathbf{y}_{A} \Big)} + \textit{CE}\Big(f(\tilde{\mathbf{x}}), \tilde{\mathbf{y}} \Big) $$ \Comment{\textit{CE} is cross-entorpy loss.}
\State $\theta_{t}=\text{SGD}\big(\theta_{t-1},\dfrac{\partial{\mathcal{L}}}{\partial{\theta_{t-1}}}\big)$

\Statex \textbf{Output:} Updated parameters $\theta_{t}$

\end{algorithmic}\label{algo}
\end{algorithm}

\begin{figure}[h!]
\begin{center}
\includegraphics[height=4.5cm,width=12.93cm]{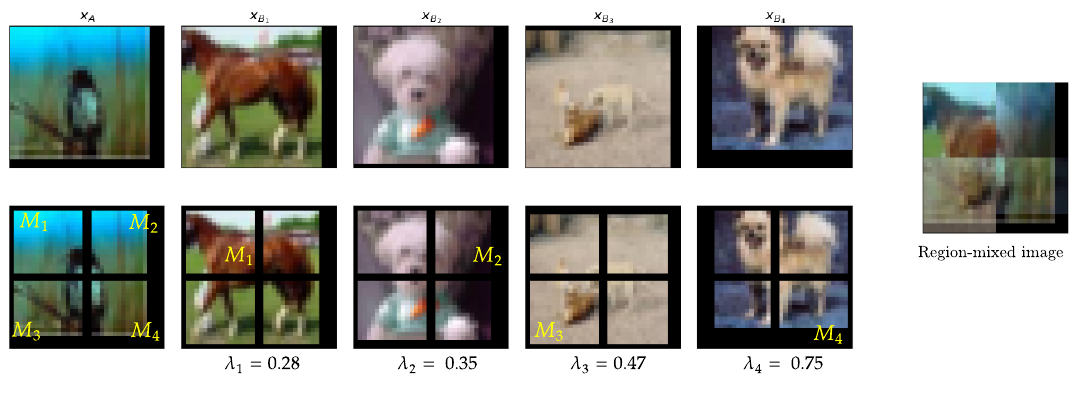}
\end{center}
\caption{Understanding region mixup}
\label{mixup_image}
\end{figure}

Although introducing stochasticity into the region selection process is an intriguing avenue for future exploration, we opted for a straightforward approach in this work.  We divide every image into non-overlapping tiles of equal size, forming regions in a grid pattern with dimensions $k\times k$ (see figure \ref{mixup_image}). In conjunction with the mixup loss, we incorporate the standard cross-entropy loss (highlighted in magenta in the algorithm \ref{algo}) for classification. Experimentally, we find  this combined loss performs better (see Table \ref{appendix-table-1} in Appendix).

\section{Experiments}
We perform image classification experiments on the CIFAR-10, CIFAR-100, and Tiny ImageNet datasets to assess the generalization capabilities of region mixup. In particular, we evaluate Mixup \citep{zhang2018mixup}, CutMix \citep{Yun2019CutMixRS} , and Region mixup for the  PreAct ResNet-18 \citep{He2016IdentityMI}.
All models undergo training on a single Nvidia RTX A5000 using PyTorch Lightning \citep{Falcon_PyTorch_Lightning_2019} for 400 epochs on the training set, employing 128 examples per minibatch. Evaluation is carried out on the test set. The training utilizes SGD with momentum, a weight decay of 0.0005, and a step-wise learning rate decay. The learning rates commence at 0.1 and undergo division by 10 after 100 and 150 epochs during the training process. We do not use dropout in these experiments.
For  all three dataset, each image is zero-padded with
two pixels on each side.
Subsequently, for CIFAR-10 and CIFAR-100, the resulting image is  cropped randomly to generate a new 32 × 32
image. For Tiny ImageNet, the random cropping process generates a new 64 × 64 image.
Next, we flip the image horizontally with a probability of 50$\%$.  
We summarize our results in Table \ref{result-table} and Table \ref{adversarial-table} (in Appendix). The results are averaged over 3 runs.


\begin{table}[ht!]
\caption{Test accuracy for the CIFAR and Tiny ImageNet experiments.}
\label{result-table}
\begin{center}
\begin{tabular}{llcccc} 
\multicolumn{1}{c}{\bf Dataset}  &\multicolumn{1}{c}{\bf Model} & \multicolumn{1}{c}{\bf CutMix} & \multirow{2}{*}{}\bf Vanilla
Mixup & \multirow{2}{*}{}\bf Region Mixup
\\  
& & & $k=1$ & $k=2$ 
\\
\hline 
CIFAR-10 & \multirow{3}{*}{ PreAct  ResNet-18}  & 95.82$\pm$0.19 & 95.89$\pm$0.12 & \textbf{96.19$\pm$0.05}
\\
CIFAR-100 & 
& \textbf{79.03$\pm0.30$}       
& 78.1 $\pm 0.60$& 78.75$\pm$ 0.28 
\\
Tiny ImageNet & & 65.76$\pm0.12$
& 65.45$\pm$0.42 
& \textbf{66.16$\pm$0.50}  
\\

\hline
\end{tabular}
\end{center}
\end{table}
\section{Discussion}
We have introduced region mixup, a simple extension of the mixup data augmentation principle. Integrating region mixup into existing mixup training pipelines requires just a few lines of code and adds minimal to no computational overhead. Through empirical findings, we observe the effectiveness of region mixup in visual recognition. We anticipate that Region Mixup will receive extensive investigation and further extensions, potentially becoming a valuable regularization tool for practitioners in deep learning.

\newpage

\section*{Acknowledgements}
This research is partially supported by
the Indo-French Centre for the Promotion of Advanced Research (IFCPAR/CEFIPRA) through Project No. 6702-2.

\section*{URM Statement}
The authors acknowledge that at least one key author of this work meets the URM criteria of the ICLR 2024 Tiny Papers Track.

\bibliography{iclr2023_conference_tinypaper}
\bibliographystyle{iclr2023_conference_tinypaper}
\newpage
\appendix
\section{Appendix}
\subsection{Ablation study}

\begin{table}[ht!]
\caption{Test accuracy for the CIFAR-100 experiment with and without the standard cross-entropy (CE) loss (highlighted in magenta in the algorithm \ref{algo}).}
\label{appendix-table-1}
\begin{center}
\begin{tabular}{llcccc}
\multicolumn{1}{c}{\bf Dataset}    & \multicolumn{1}{c}{\bf CutMix} & \multirow{2}{*}{}\bf Vanilla
Mixup & \multirow{2}{*}{}\bf Region Mixup
\\  
 & & $k=1$ & $k=2$ 
\\
\hline 
w/o  standard CE
   &   $78.27\pm0.34$ 
   & $77.5\pm0.22$ 
   & $76.36\pm0.26$ 
\\
with standard CE
& \textbf{$79.03\pm0.30$}         
& 78.1 $\pm 0.60$& 78.75$\pm$ 0.28 
\\
\hline

\end{tabular}
\end{center}
\end{table}

\subsection{Adversarial robustness}

We assess the robustness of the trained
models against adversarial samples. Adversarial examples are generated (in one single step) using the Fast Gradient Sign Method (FGSM) \citep{DBLP:journals/corr/GoodfellowSS14}, with the assumption that the adversary possesses complete information about the models, thereby conducting a white-box attack. Following \cite{zhang2018mixup}, we constrain our experiment to basic FGSM attacks as the strength of iterative PGD attacks diminishes the practical relevance of any observed performance enhancements.  For the black-box attack setting,  we consider $l_{\infty}$-square attack \citep{ACFH2020square} with a constraint on the query budget limited to 100 queries.
We use torchattack \citep{TorchAttack} to
launch these attacks.
We report test accuracies  after the attack in Table \ref{adversarial-table} and Table \ref{adversarial-table-blackbox}.

\begin{table}[ht!]
\caption{Test accuracy on white-box FGSM adversarial examples.}
\label{adversarial-table}
\begin{center}
\begin{tabular}{llcccc} 
\multicolumn{1}{c}{\bf Dataset}  &\multicolumn{1}{c}{\bf Model} & \multicolumn{1}{c}{\bf CutMix} & \multirow{2}{*}{}\bf Vanilla
Mixup & \multirow{2}{*}{}\bf Region Mixup
\\  
& & & $k=1$ & $k=2$ 
\\
\hline 
CIFAR-10 & \multirow{3}{*}{ PreAct  ResNet-18}  &  36.15$\pm$2.73 
& 52.40$\pm$6.60 
& \textbf{58.96$\pm$5.23} 
\\
CIFAR-100 & 
&  13.08$\pm$0.87 
& 18.56$\pm$1.22  
& \textbf{22.77$\pm$1.20}  
\\
Tiny ImageNet & 
& 	\textbf{2.69 $\pm$ 0.19} 
& 1.71$\pm$0.18 
& 2.20$\pm$0.22      
\\

\hline
\end{tabular}
\end{center}
\end{table}


\begin{table}[ht!]
\caption{Test accuracy on black-box Square Attack ($l_{\infty}$). The black-box attacks are provided with a budget of 100 queries}
\label{adversarial-table-blackbox}
\begin{center}
\begin{tabular}{llcccc} 
\multicolumn{1}{c}{\bf Dataset}  &\multicolumn{1}{c}{\bf Model} & \multicolumn{1}{c}{\bf CutMix} & \multirow{2}{*}{}\bf Vanilla
Mixup & \multirow{2}{*}{}\bf Region Mixup
\\  
& & & $k=1$ & $k=2$ 
\\
\hline 
CIFAR-10 & \multirow{3}{*}{ PreAct  ResNet-18}  &  31.76$\pm$1.70 
& \textbf{52.18$\pm$1.18} 
& 51.35$\pm$1.99 
\\
CIFAR-100 & 
& 10.54$\pm$0.59 
& \textbf{20.03$\pm$0.46} 
& 18.70$\pm$1.05 
\\
Tiny ImageNet & 
& 		17.03$\pm$0.18 
& 24.78$\pm$0.19 
& \textbf{25.02$\pm$0.69} 
\\

\hline
\end{tabular}
\end{center}
\end{table}

\newpage
\subsection{class activation mappings}
We qualitatively compare Mixup, CutMix, and Region Mixup  using  class activation mappings (CAM) generated by Grad-CAM++ \citep{gradcam++} on Tiny ImageNet dataset. We use the final residual block (layer4) of PreAct ResNet as the target layer to compute CAM.

\begin{figure}[h!]
\begin{center}
\includegraphics[width=14cm,height=12cm]{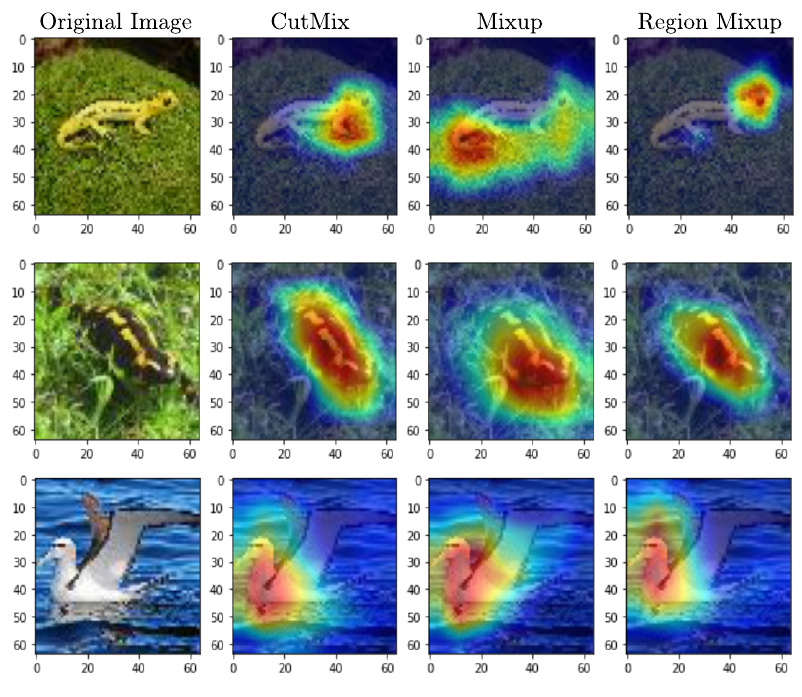}
\end{center}
\caption{ 
Class activation mapping (CAM) \citep{Zhou2015LearningDF} visualizations on Tiny ImageNet
using Grad-CAM++ \citep{gradcam++}.
}
\label{saliency_image}
\end{figure}

\end{document}

%% file: math_commands.tex
\usepackage{amsmath,amsfonts,bm}

\def\eqref#1{equation~\ref{#1}}

\def\1{\bm{1}}

\def\ra{{\textnormal{a}}}

\def\rx{{\textnormal{x}}}

\def\rva{{\mathbf{a}}}

\def\erva{{\textnormal{a}}}

\def\vmu{{\bm{\mu}}}
\def\vtheta{{\bm{\theta}}}
\def\va{{\bm{a}}}

\def\vx{{\bm{x}}}

\def\eva{{a}}

\def\mA{{\bm{A}}}

\def\mH{{\bm{H}}}

\def\mJ{{\bm{J}}}

\def\mX{{\bm{X}}}

\def\mSigma{{\bm{\Sigma}}}

\DeclareMathAlphabet{\mathsfit}{\encodingdefault}{\sfdefault}{m}{sl}
\SetMathAlphabet{\mathsfit}{bold}{\encodingdefault}{\sfdefault}{bx}{n}
\newcommand{\tens}[1]{\bm{\mathsfit{#1}}}
\def\tA{{\tens{A}}}

\def\tX{{\tens{X}}}

\def\sA{{\mathbb{A}}}
\def\sB{{\mathbb{B}}}

\def\sS{{\mathbb{S}}}

\def\emA{{A}}

\newcommand{\etens}[1]{\mathsfit{#1}}

\def\etA{{\etens{A}}}

\newcommand{\E}{\mathbb{E}}

\newcommand{\R}{\mathbb{R}}

\newcommand{\KL}{D_{\mathrm{KL}}}
\newcommand{\Var}{\mathrm{Var}}

\newcommand{\Cov}{\mathrm{Cov}}

\newcommand{\normltwo}{L^2}
\newcommand{\normlp}{L^p}
